\documentclass[wcp]{jmlr_arxiv}
\usepackage{longtable}% for long tables

\usepackage{booktabs}
\usepackage{bm}
\usepackage{ascmac}
\usepackage{algorithm}
\usepackage{algorithmic}

\newcommand{\pdif}[2]{\frac{\partial #1}{\partial #2}}
\newcommand{\squarematrixtwo}[4]{\left( \begin{array}{cc} #1 & #2\\ #3 & #4\end{array} \right)}
\newcommand{\vectortwo}[2]{\left( \begin{array}{c} #1\\ #2\end{array}\right)}
\newcommand{\heikin}[1]{\langle #1 \rangle}
\newcommand{\kitai}{\mathbb{E}}
\newcommand{\zissu}{\mathbb{R}}

\newcommand{\T}{\top}
\newcommand{\E}{\mathbb E}
\DeclareMathOperator*{\argmin}{arg\,min}
\DeclareMathOperator*{\argmax}{arg\,max}

\title[Bayesian Masking: Sparse Bayesian Estimation with Weaker Shrinkage Bias]{Bayesian Masking: Sparse Bayesian Estimation with Weaker Shrinkage Bias}

  \author{\Name{Yohei Kondo} \Email{ykondo@sys.i.kyoto-u.ac.jp}\\
  \Name{Shin-ichi Maeda} \Email{ichi@sys.i.kyoto-u.ac.jp}\\
  \addr Graduate School of Informatics, Kyoto University, Kyoto, Japan
  \AND
  \Name{Kohei Hayashi} \Email{hayashi.kohei@gmail.com}\\
  \addr National Institute of Informatics, Tokyo, Japan\\
  \addr Kawarabayashi Large Graph Project, ERATO, JST
 }

\begin{document}

\maketitle

\begin{abstract}
 A common strategy for sparse linear regression is to introduce regularization,
 which eliminates irrelevant features by letting the
  corresponding weights be zeros.
  However, regularization often shrinks the estimator for
  relevant features, which leads to incorrect feature selection.
  
  Motivated by the above-mentioned issue, we propose Bayesian masking (BM), a sparse estimation method which
  imposes no regularization on the weights. 
  The key concept of BM is to introduce binary latent variables that randomly mask features.
  Estimating the masking rates determines the relevance of the features automatically.
  We derive a variational Bayesian inference algorithm that maximizes the lower bound of
  the factorized information criterion (FIC), which is a recently developed asymptotic criterion for evaluating the marginal log-likelihood.
  In addition, we propose reparametrization to accelerate the convergence of the derived algorithm.
  Finally, we show that BM outperforms Lasso and automatic relevance determination (ARD) in terms of the sparsity-shrinkage trade-off. 
\end{abstract}

%%% Local Variables:
%%% mode: latex
%%% TeX-master: "acml15_ard.tex"
%%% End:

\begin{keywords}
Sparse estimation, Factorized information criterion, Lasso, Automatic relevance determination
\end{keywords}

\section{Introduction}
In sparse linear regression, various approaches impose sparsity by
implementing regularization on a weight parameter. For example,
Lasso~\citep{tibshirani94} introduces sparsity by regularizing the
weights by L1 norm. Automatic relevance determination
(ARD)~\citep{mackay94,neal96} regularizes the weights by a prior
distribution, with hyperparameters indicating the relevance of the input
features. Empirical Bayes estimation of the hyperparameters thus
eliminates irrelevant features automatically. Although ARD is
notorious for its slow convergence, several authors have improved the algorithm
(e.g., \citep{wipf08}).

The trade-off between sparsity and shrinkage is a crucial issue in
sparse regularization methods~\citep{aravkin14}. In Lasso, for
example, a large regularization constant incorporates strong
sparsity and is more likely to estimate the weights of irrelevant
features as zero, which is desirable in terms of
interpretability. However, it also shrinks the weights of relevant
features and may eliminate them. 
ARD suffers from the same problem, although the
bias of ARD is weaker than that of Lasso~\citep{aravkin14}.
Because both sparsity and shrinkage are caused by regularization, the
shrinkage effects are inevitable as long as we use the regularization
scheme.

To address this issue, we propose an alternative method for sparse estimation, namely, Bayesian masking (BM),
 which differs from existing methods in that it does not impose any regularization on the weights.
Our contributions can be summarized as follows.
\begin{itemize}
\item The BM model
  (Section~\ref{sec:bayes-sparse-line}).
  The BM model introduces binary latent variables into a linear regression model.
  The latent variables randomly mask features to be zero at each sample according to the priors that are defined for each feature, but shared among samples.
  The estimation of the priors on the masking rates determines the relevance of the features.
\item A variational Bayesian inference algorithm for the BM model (Section~\ref{sec:algorithm}).
The EM-like coordinate ascent algorithm maximizes the lower bound of the factorized information criterion (FIC).
The convergence of the algorithm is accelerated by combining gradient ascent and reparametrization (Section~\ref{subsec:reparametrization}),
which are motivated by previous studies on convergence analysis of coordinate ascent and information geometry. 
\item Analytic forms of the one-dimensional (1D) estimators of Lasso, ARD
  (Section~\ref{sec:shrink-bias:-tent}), and BM
  (Section~\ref{sec:fab-estimator}). The analytic estimators of these methods provide
  insights into their shrinkage mechanisms.
\end{itemize}
Through numerical experiments, we empirically show that the proposed method outperforms Lasso and ARD in terms of the sparsity-shrinkage trade-off.

\subsection{Notation}

Hereafter, $\bm{x}_{n}$ denotes a column vector of the $n$-th row of a matrix $X$.

%%% Local Variables:
%%% mode: latex
%%% TeX-master: "acml15_ard.tex"
%%% End:

\section{Background}\label{sec:problem-setting}

\subsection{Linear Regression and Least Squares}

Consider a linear regression model:
\begin{eqnarray}
	\bm{y} = X \bm{\beta} + \bm{\epsilon},\label{eq:problem}
\end{eqnarray}
where $\bm{y} \in \zissu^N$ is a vector of target values, $X \in
\zissu^{N\times K}$ is a matrix of explanatory variables, $\bm{\beta}
\in \zissu^K$ is a vector of weight parameters, and
$\bm{\epsilon} \sim N(0,\lambda^{-1} I)$ denotes observation noise.
Further, $N$ is the number of samples and $K$ is the number of features. Because the noise is i.i.d. Gaussian, the maximum
likelihood estimator (MLE) is given as the least-squares (LS) solution:
\begin{align}\label{eq:LS}
  \hat{\bm{\beta}}_{\rm{LS}} 
&= \argmin_{\bm{\beta}} \frac{\lambda}{2}\| \bm{y} - X \bm{\beta}\|^2_2
=(X^\T X)^{-1}X^\T\bm y.
\end{align}

\subsection{Lasso}

Lasso is formulated as the L1-penalized regression problem, and the
estimator is given as
\begin{eqnarray}\label{eq:lasso}
	\hat{\bm{\beta}}_{\rm{Lasso}} = \argmin_{\bm{\beta}} \frac{\lambda}{2}\| \bm{y} - X \bm{\beta}\|^2_2  + \alpha \|\bm{\beta}\|_1,
\end{eqnarray}
where $\alpha (> 0)$ is a regularization constant. 

\subsection{ARD}

Consider the prior distribution of $\bm{\beta}$:
\begin{eqnarray}
	p(\bm{\beta} | \bm{\gamma}) = N(\bm{\beta}| 0, \Gamma),
\end{eqnarray}
where $\Gamma=\mbox{diag}(\bm{\gamma})$ and $\bm{\gamma}$ is the
hyperparameter determining the variance of the prior. ARD determines
$\bm{\gamma}$ by the empirical Bayes principle, i.e., by maximizing the
marginal log-likelihood: $\hat{\bm{\gamma}}=\argmax_\gamma\int
p(\bm{y}|\bm{\beta})p(\bm{\beta}|\bm{\gamma},\lambda)
\mathrm{d}\bm{\beta}$.  Then, the estimator of $\bm{\beta}$ is then often
given by the posterior mean with plugged-in $\hat{\bm{\gamma}}$~\citep{wipf08,aravkin14}:
\begin{eqnarray}\label{eq:ard}
  \hat{\bm{\beta}}_{\rm{ARD}} = \hat\Gamma X^\T(\lambda^{-1} I + X\hat\Gamma X^\T)^{-1}\bm{y}.
\end{eqnarray}
Clearly, for $\lambda^{-1}>0$, $\hat\gamma_k = 0$ results in $\beta_k = 0$ for any $k$.

%%% Local Variables:
%%% mode: latex
%%% TeX-master: "acml15_ard_debug.tex"
%%% End:

\section{Trade-off between Sparsity and Shrinkage}\label{sec:shrink-bias:-tent}

Concerning the trade-off between sparsity and shrinkage, \cite{aravkin14}
derived the upper bounds of the estimators for $K=2$ and showed that
undesirable shrinkage occurs for both Lasso and ARD; specifically, the
shrinkage bias of Lasso is larger than that of ARD when an unnecessary feature is correctly pruned.

In this section, we revisit the above-mentioned issue. To understand how sparse
regularization works, we derive the exact forms of the estimators for
$K=1$ and show that ARD is better than Lasso in terms of the
sparsity-shrinkage trade-off. Although our analysis is much simpler
than the earlier study by \cite{aravkin14}, it is meaningful
because our derived estimators
\begin{itemize}
\item are exact and analytically written (no approximation is needed) and
\item highlight the significant differences between ARD and Lasso.
\end{itemize}

\subsection{1D Estimators}

\paragraph{LS}

When $K=1$, the matrix inverse in Eq.~\eqref{eq:LS} becomes a scalar
inverse and $\hat{\beta}_{\rm{LS}}$ is simply written as
\begin{align}
  \hat{\beta}_{\rm{LS}}=\frac{\bm{x}^\T\bm{y}}{\bm{x}^\T\bm{x}},
\end{align}
where we let $\bm{x}$ represent $X$ to emphasize the dimensionality.

\paragraph{Lasso}

For $\beta\geq 0$, the L1-penalty becomes $\alpha\beta$. Thus, Eq.~\eqref{eq:lasso} yields a stationary point
$\hat{\beta}_{\rm{LS}}-\alpha/\bm{x}^\T\bm{x}$ where
$\hat{\beta}_{\rm{LS}}=\bm{x}^\T\bm{y}/\bm{x}^\T\bm{x}$. For
$\beta<0$, the solution is the same except that the sign is
reversed. Combining both cases yields the solution for all $\beta$:
\begin{align}\label{eq:lasso-1d}
  \hat{\beta}_{\rm{Lasso}}=\mathrm{sign}(\hat{\beta}_{\rm{LS}})\max(0,
  |\hat{\beta}_{\rm{LS}}|
    -\frac{2\alpha}{\lambda\bm{x}^\T\bm{x}}).
\end{align}

\paragraph{ARD}

Similar to Lasso, the 1D estimator of ARD is analytically
written as\footnote{The full derivation of Eq.~\eqref{eq:ard-1d} is
  shown in Appendix~\ref{sec:ard-estimator-when}.}
\begin{align}\label{eq:ard-1d}
  \hat{\beta}_{\rm{ARD}} =
  \mathrm{sign}(\hat{\beta}_{\rm{LS}})\max \bigl(0, 
  |\hat{\beta}_{\rm{LS}}| - \frac{1}{\lambda|\bm{x}^\T\bm{y}|} \bigr).
\end{align}
Note that we assumed that $\lambda$ is known.\footnote{Practically, $\lambda$ is set
to the unbiased version of MLE~\citep{aravkin14}.}

\subsection{Comparison of LS, Lasso and ARD}

Although their regularizations are different, Lasso and ARD
have the same shrinkage mechanism --- subtracting the constant from
$\hat{\beta}_{\rm{LS}}$ and cropping $\hat{\beta}_{\rm{LS}}$ to zero if its magnitude
 is smaller than that of the constant. Since the
constants in Eqs.~\eqref{eq:lasso-1d} and \eqref{eq:ard-1d} are both larger than
zero except for the noiseless case (i.e., $\lambda=\infty$), the
shrinkage bias is inevitable in both Lasso with $\alpha>0$ and ARD.
On the other hand, this retraction to zero is necessary for sparsity because it prunes the irrelevant features.

\begin{figure}[tb]
  \begin{center}
  \includegraphics[width=0.7\textwidth,trim=0 0 0 100,clip]{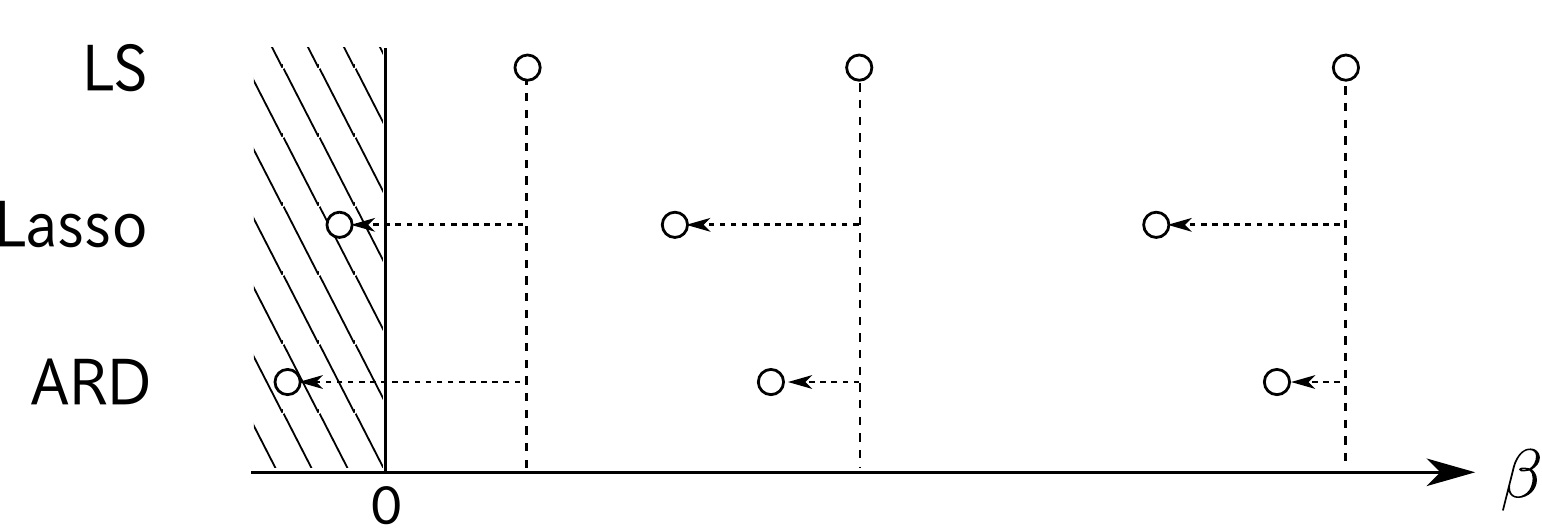}
  \caption{Illustrative comparison of the shrinkage effect with
    Lasso and ARD in the 1D case. Lasso shifts the
    estimator to zero by the same value, whereas the shrinkage of ARD is larger
    when the LS estimator is smaller.}
  \label{fig:shrinkage}
  \end{center}
\end{figure}

It is worth noting that the bias of ARD is much weaker than that of Lasso when the scale of $\hat{\beta}_{\rm{ARD}}$ is large. This
is easily confirmed by transforming the constant in
Eq.~\eqref{eq:ard-1d} as $(\lambda|\bm{x}^\T\bm{y}|)^{-1} =
(\lambda\bm{x}^\T\bm{x}|\hat{\beta}_{\rm{LS}}|)^{-1}$, which indicates that
the shrinkage is weak when $|\hat{\beta}_{\rm{LS}}|$ is large but
strong when $|\hat{\beta}_{\rm{LS}}|$ is close to zero. Compared to
Lasso, this behavior of ARD is desirable, as it maintains
sparsity for weak features while alleviating unnecessary
shrinkage. Figure~\ref{fig:shrinkage} shows how shrinkage occurs
in Lasso and ARD.

%%% Local Variables:
%%% mode: latex
%%% TeX-master: "acml15_ard_debug.tex"
%%% End:

\section{BM Model and Inference Algorithms}\label{sec:bayes-sparse-line}

\subsection{BM Model}\label{sec:bm_model}

Obviously, the shrinkage bias of Lasso and ARD comes from their imposition regularization on the weights.
For example, if $\alpha=0$ in Lasso, the loss function becomes equivalent to that of LS, and of course, no shrinkage occurs. Using $\bm{\gamma}=\infty$ yields the same result in ARD.

Hence, we introduce a new estimation method that maintains sparsity by using latent variables instead of regularization.
Let $Z \in \{0, 1\}^{N \times K}$ be binary latent
variables having the same dimensionality of $X$. We insert $Z$ between
$X$ and $\bm{\beta}$, i.e.,
\begin{eqnarray}\label{eq:model}
	\bm{y} = (X \circ Z) \bm{\beta} + \bm{\epsilon},
      \end{eqnarray}
where '$\circ$' denotes the Hadamard product (i.e., element-wise
multiplication). 

$Z$ masks the features randomly at each sample.
For Bayesian treatment, we introduce prior distributions. We assume
that $Z$ follows a Bernoulli prior distribution as $z_{nk} \sim \mbox{Bern} (\pi_k)$, where $\pi_k$ indicates how feature $k$ is relevant.
Then, estimating the priors on the masking rates automatically determines the relevance of the features.

We also introduce the priors for $\bm{\beta}$ and $\lambda$; however,
we set them to be as weak as possible so that their effects are negligible when $N$ is sufficiently large. Thus,
we simply employ them as constants as they do not depend on $N$, i.e., $\log p(\bm{\beta},\lambda)=O(1)$.

\subsection{FAB-EM Algorithm}\label{sec:algorithm}

Our approach is based on the concept of Bayesian model selection;
the central task is to evaluate the marginal log-likelihood.
However, in our case, the marginal likelihood is intractable.

Thus, we adopt FIC, a recently proposed approximation for the marginal log-likelihood~\citep{fujimaki12a,hayashi13,hayashi15}.  
We also adopt the factorized asymptotic Bayesian inference (FAB),
which provides a tractable algorithm for parameter inference and model
pruning by optimizing the lower bound of FIC.  The FAB algorithm
alternately maximizes the lower bound in an EM-like manner.

To obtain the lower bound of FIC, we introduce a mean-field approximation on the posterior distribution of $Z$ as
$q(\bm{z}_n) = \prod_k q(z_{nk}) = \prod_k \mbox{Bern} (\mu_{nk})$. Then we obtain the
objective function as
\begin{eqnarray}
	\mathcal{G}(\{\bm{\mu}_n\}, \bm{\beta}, \lambda, \bm{\pi}) &=& \kitai_q [\log p(\bm{y} |X, Z, \bm{\beta}, \lambda)] + \kitai_q [\log p(Z | \bm{\pi})]\nonumber \\
	 && - \frac{1}{2} \sum_k \left( \log (N\pi_k) + \frac{\sum_n \kitai_q[z_{nk}] / N - \pi_k}{\pi_k} \right) - \frac{K+1}{2} \log N\nonumber \\
	 && + \sum_{n,k} H(q(z_{nk})),\label{eq:FIClowerbound}
\end{eqnarray}
where $\kitai_q$ means the expectation under $q = q(Z)$, and $H$ is the entropy.
The derivation of Eq.~(\ref{eq:FIClowerbound}) is described in Appendix~\ref{sec:derivation_lowerbound}.

\paragraph{FAB E-step}

In the FAB E-step, we update $\{\mu_{nk}\}$. By taking the gradient of $\mathcal{G}$ w.r.t. $\mu_{nk}$ and setting it to zero, we obtain the following
fixed-point equations:
\begin{eqnarray}
	\mu_{nk} = \sigma \left( c_{nk} + \log \frac{\pi_k}{1 - \pi_k} - \frac{1}{2N\pi_k} \right)\label{eq:FAB_Estep}
\end{eqnarray}
where $\sigma(\cdot)$ is the sigmoid function and
$c_{nk} = x_{nk} \beta_k \lambda (y_n - \frac{1}{2} x_{nk} \beta_k - \sum_{l \neq k} \mu_{nl} x_{nl} \beta_l)$.
Updating Eq.~(\ref{eq:FAB_Estep}) several times give us $\{\mu_{nk}\}$ at a local maximum of $\mathcal{G}$.

\paragraph{FAB M-step}

Since only the first and second terms in Eq.~\eqref{eq:FIClowerbound} are relevant, the FAB M-step is equivalent to the M-step in the EM algorithm.
We have the closed-form solutions to update $\bm{\beta}, \lambda$, and $\bm{\pi}$ as
\begin{eqnarray}
	\bm{\beta} &=& \Omega^{-1} (X \circ M)^\top \bm{y},\label{eq:FAB_Mstep1}\\
	\frac{1}{\lambda} &=&  \frac{\sum_n (y_{n}^2 - 2y_{n} (\bm{x}_n \circ \bm{\mu}_n)^\top \bm{\beta} + (\bm{x}_n \circ \bm{\beta})^\top \kitai_q [\bm{z}_n \bm{z}_n^\top] (\bm{x}_n \circ \bm{\beta}))}{N} \label{eq:FAB_Mstep2}\\
	\pi_k &=& \sum_n \frac{\mu_{nk}}{N},\label{eq:FAB_Mstep3}
\end{eqnarray}
where $\Omega= \sum_n (\bm{x}_n \bm{x}_n^\top) \circ \kitai_q [\bm{z}_n \bm{z}_n^\top]$ and $M = (\bm{\mu}_1, ..., \bm{\mu}_N)^\top$.
We note that $\kitai_q [\bm{z}_n \bm{z}_n^T] = \bm{\mu}_n \bm{\mu}_n^T + \mbox{diag}(\bm{\mu}_n  - \bm{\mu}_n \circ \bm{\mu}_n)$.

\paragraph{Pruning step}

As noted in previous papers, the third term in $\mathcal{G}$ penalizes $\{\mu_{nk}\}$ and automatically eliminates irrelevant features.
Then, when $\pi_k = \sum_n \mu_{nk}/N < \delta$, we remove the corresponding feature from the model.
The pseudo-code of the resulting algorithm is given in Algorithm~\ref{alg:FAB}.

\begin{algorithm}
\caption{Bayesian masking by FAB-EM algorithm}
\label{alg:FAB}
\begin{algorithmic}[1]
\STATE Initialize $(\{\bm{\mu}_n\}, \bm{\beta}, \lambda, \bm{\pi})$
\REPEAT
\STATE Update $\{\mu_{nk}\}$ by Eq.~(\ref{eq:FAB_Estep})
\FOR{$k = 1, ..., K$}
\IF{$\sum_n \mu_{nk}/N < \delta$}
\STATE Remove $k$-th dimension from the model
\ENDIF
\ENDFOR
\STATE Update $(\bm{\beta}, \lambda, \bm{\pi})$ by Eqs.~(\ref{eq:FAB_Mstep1}-\ref{eq:FAB_Mstep3})
\UNTIL{termination criterion is met}
\end{algorithmic}
\end{algorithm}

\subsection{Analysis of FAB Estimator}\label{sec:fab-estimator}

As in Section~\ref{sec:shrink-bias:-tent}, we investigate the FAB estimator of $\bm{\beta}$. If $\bm{y}$ follows the linear
model~\eqref{eq:problem} with $\bm{\beta}=\bm{\beta}^*$, the FAB estimator
is expectedly obtained as
\begin{align}
  \E_{\bm\epsilon}[\hat{\bm{\beta}}_{\rm{FAB}}]
  &=
  \Omega^{-1}\tilde{X}^\T X\bm{\beta}^* \notag
  \\
  &=\bm{\beta}^* + \Omega^{-1}\bm{b}\label{eq:fab-bias}
\end{align}
for any $q(Z)$, where $\tilde{X}=X\circ M$ and $b_k =
(\bm{x}_k\circ\bm{\mu}_k)^\T\sum_{l\not=k}\beta^*_l(\bm{x}_l\circ(\bm{1}-\bm{\mu}_l))$.

Eq.~\eqref{eq:fab-bias} immediately suggests that
$\hat{\bm{\beta}}_{\rm{FAB}}$ is biased by $\Omega^{-1}\bm{b}$. The bias
consists of the two cross terms: $(\bm{x}_k,\bm{x}_l)$ and
$(\bm{\mu}_k,\bm{1}-\bm{\mu}_l)$. Thus, the bias increases when $\bm{x}_k$ and
$\bm{x}_l$ are correlated and $\bm{z}_k$ and $\bm{z}_l$ are negatively
correlated. On the other hand, the cross terms become zero when
$\mu_{nk}=0$ or $\mu_{nl}=1$ for all $n$. This also implies that the
bias is weakened by an appropriately estimated $q$. In
Section~\ref{sec:simple-example}, we numerically evaluate the bias of
FAB with Lasso and ARD, showing that FAB achieves the lowest bias.

Remarkably, when $K=1$, no cross term appears and the bias
vanishes for any $q$ satisfying $\pi>0$. Furthermore, the 1D estimator is simply written as
\begin{align}
  \hat{\beta}_{\rm{FAB}}=
\frac{\tilde{\bm{x}}^\T\bm{y}}{\tilde{\bm{x}}^\T\bm{x}}.
\end{align}
Again, if $\mu_{nl}=1$ for all $n$, $\hat{\beta}_{\rm{FAB}}$ recovers
$\hat{\beta}_{\rm{LS}}$.

\subsection{FAB-EG and Hybrid FAB-EM-EG Algorithms}\label{subsec:reparametrization}

\cite{hayashi13} reported that model pruning in FAB is slow, and we find that our algorithm suffers from the same drawback.
In our case, \{$\pi_k$\} for irrelevant features requires many iterations until convergence to zero
although the weights $\bm{\beta}$  and $\{\pi_k\}$ for relevant features converge rapidly.
To overcome the problem of slow convergence, we combine gradient ascent and reparametrization as discussed below.

First, we replace the FAB-M step by gradient ascent, which is motivated by previous studies~\citep{salakhutdinov03,maeda07}
on convergence analysis of the EM algorithm.
Thus, we find that gradient ascent certainly helps; however, the convergence remains slow.
This is because the model distribution is insensitive to the direction of $\pi_k$ when $\beta_k$ is small.
This means that the gradient for $\pi_k$ would be shallow for an irrelevant feature $k$, since the estimator of $\beta_k$ takes a small value for the feature.

The natural gradient (NG) method \citep{amari98} can effectively overcome the insensitivity in the model distribution.
The key concept of the NG method is to adopt the Fisher information matrix as a metric on a parameter space (the so-called the Fisher information metric)
in order to define the distance by the Kullback-Leibler (KL) divergence between model distributions
instead of the Euclidean distance between parameter vectors.
Thus, parameter updates by the NG method can make steady changes in the model distribution at each iteration.
Amari and his co-workers showed that the NG method is especially effective when the parameter space contains singular regions, i.e.,
 regions where the Fisher information matrix degenerates~\citep{amari06,wei08}.
 This is because the problem of shallow gradient is severe around a singular region, since gradient completely vanishes along with the region.
 Hence, learning by ordinary gradient is often trapped around singular regions and remains trapped for many iterations, even when the models in the regions show poor agreement with data.
  In contrast, learning by the NG method is free from such a slow down.
  
In our case, the model has two singular regions for each feature, i.e., $\beta_k = 0$ and $\pi_k = 0$.
In particular, singular region $\beta_k = 0$ causes the slow convergence.
Hence, the NG method should be effective; however, the evaluation of the Fisher information metric is computationally expensive in our case.
Therefore, as an alternative, we propose a simple reparametrization that approximates the Fisher information metric.
Our strategy is to perform a block diagonal approximation of the full matrix, as has been performed recently in \citep{desjardins15} for neural networks.

Toward this end, we examine the Fisher information metric in the single-parameter case ($K = 1$), which approximates diagonal blocks of the full metric.
Then, the model of interest here is simply $y_n \sim \pi N(y_n | x_n \beta, \lambda^{-1}) + (1-\pi) N(y_n | 0, \lambda^{-1})$.
We focus on the case of small $\beta$ because the slow learning of $\pi$ is prominent in this region as noted above.
Although the exact form of the metric is difficult to compute, our focus on the small-$\beta$ case allows further approximation.
Taylor expansion around $\beta = 0$ and lowest-order approximation give us a metric tensor:
\begin{eqnarray}
	G \equiv \squarematrixtwo{G_{\beta \beta}}{G_{\beta \pi}}{G_{\beta \pi}}{G_{\pi \pi}} 
	=\lambda N \heikin{x^2} \squarematrixtwo{ \lambda \beta^2 N \heikin{x^2} f(\pi) + \pi^2}{\beta \pi}{\beta \pi}{\beta^2}.\label{eq:Fisher_metric},
\end{eqnarray}
where $f(\pi)$ represent polynomials of $\pi$ and $\heikin{x^2} \equiv \sum_n x_n^2/ N$.
The key factor in $G$ is $G_{\pi \pi} \propto \beta^2$ because this represents the fact that a smaller value of $\beta$ makes the model more insensitive to changes in $\pi$. 

The approximated metric obtained above is complicated and difficult to handle.
Hence, we consider the following simple reparametrization for $k = 1,..., K$:
\begin{eqnarray}
	(\beta_k, \pi_k) \to (\beta_k, s_k = \beta_k \pi_k).
\end{eqnarray}
Let us show what metric is introduced in $(\beta, \pi)$-space through the reparametrization.
Toward this end, we recall that considering usual gradient ascent on ($\beta, s$)-space corresponds to introducing a metric $J^\top J$ in the original space,
where $J$ is the Jacobian matrix for the reparametrization.
Then, the reparametrization corresponds to introducing a block diagonal metric tensor in which each diagonal block is given by
\begin{eqnarray}
	G' = \squarematrixtwo{1 + \pi_k^2}{\beta_k \pi_k}{\beta_k \pi_k}{\beta_k^2}. \label{eq:our_metric}
\end{eqnarray}
The similarity between $G$ and $G'$ is clear. Although they are not identical, $G'$ is simpler and shares the key factor as $G'_{\pi \pi} \propto \beta^2$.

\paragraph{FAB-G step}
In the FAB-G step, we replace the FAB-M step by gradient ascent.
We calculate the gradient on the parameter space after reparametrization ($\beta_k, s_k$), and project it onto the original space.
Then, we obtain the following update rule for ($\beta_k, \pi_k$):
\begin{eqnarray}
	\renewcommand{\arraystretch}{1.5}
	\vectortwo{\beta_k^{t+1}}{\pi_k^{t+1}} = \vectortwo{\beta_k^{t}}{\pi_k^{t}} + \eta_t \squarematrixtwo{1}{-\frac{\pi_k}{\beta_k}}{-\frac{\pi_k}{\beta_k}}{\frac{1+\pi_k^2}{\beta_k^2}} \vectortwo{\pdif{\mathcal{G}}{\beta_k}}{\pdif{\mathcal{G}}{\pi_k}}, \label{eq:FAB_Gstep}
	\renewcommand{\arraystretch}{1}
\end{eqnarray}
where $t$ is the number iterations and $\{\eta_t\}$ are the learning coefficients.
We note that the update of $\pi_k$ by $\pdif{\mathcal{G}}{\pi_k}$ is scaled by $\beta_k^{-2}$, and then accelerated when $\beta_k$ is small, as expected.
When $\pi_k = 1$, we update only $\beta_k$ by $\eta \pdif{\mathcal{G}}{\beta_k}$.
Since the singularity problem comes from $\bm{\beta}$ and $\bm{\pi}$, not $\lambda$, updating $\lambda$ retains the closed-form solution Eq.~(\ref{eq:FAB_Mstep3}).

We refer to the FAB algorithm as the FAB-EG algorithm when the G step replaces the M step.
Even though the FAB-EG algorithm shows faster convergence, the fast initial progress in the FAB-EM algorithm remains an attractive feature.
Thus, to exploit both benefits, we propose a hybrid algorithm in which the learning progresses by FAB-EM initially and by FAB-EG later.
The pseudo-code of the hybrid algorithm is given in Algorithm~\ref{alg:FAB2}.

\begin{algorithm}
\caption{Bayesian masking by Hybrid FAB-EM-EG algorithm}
\label{alg:FAB2}
\begin{algorithmic}[1]
\STATE Initialize $(\{\bm{\mu}_n\}, \bm{\beta}, \lambda, \bm{\pi})$
\STATE $t \gets 0$
\REPEAT
\STATE Update $\{\bm{\mu}_n\}$ by FAB-E step
\FOR{$k = 1, ..., K$}
\IF{$\sum_n \mu_{nk}/N < \delta$}
\STATE Remove $k$-th dimension from the model
\ENDIF
\ENDFOR
\IF{$t < T$}
\STATE Update $(\bm{\beta}, \lambda, \bm{\pi})$ by FAB-M step
\ELSE
\STATE Update $(\bm{\beta}, \lambda, \bm{\pi})$ by FAB-G step
\ENDIF
\STATE $t \gets t+1$
\UNTIL{termination criterion is met}
\end{algorithmic}
\end{algorithm}

%%% Local Variables:
%%% mode: latex
%%% TeX-master: "acml15_ard_debug.tex"
%%% End:
\section{Experiments}\label{sec:experiments}

\subsection{Overview}

First, we evaluate BM (i.e., the proposed method), Lasso, and ARD with a simple example where $K=2$,
 and we show that BM outperforms the other two in terms of the sparsity-shrinkage trade-off.
Second, using the same example, we show how the parameters are updated by the FAB-EM and FAB-EG algorithms.
Specifically, we highlight how the use of gradient ascent and the introduction of the reparametrization help to avoid being trapped in the singular region.
Third, we demonstrate that the hybrid FAB-EM-EG algorithm converges faster than the FAB-EM algorithm.
Finally, we evaluate BM, Lasso, and ARD again using larger values of $K$.

\subsection{Experiment with Two Features}\label{sec:simple-example}

For demonstration purposes, we borrowed a simple $K=2$ setting from \cite{aravkin14} who considered that
\begin{eqnarray}
	\vectortwo{y_1}{y_2} = \squarematrixtwo{1}{0}{0.5}{1} \vectortwo{\beta_1}{\beta_2} + \vectortwo{\epsilon_1}{\epsilon_2},\label{eq:toy_model}
\end{eqnarray}
where $\epsilon_1$ and $\epsilon_2$ are sampled from $N(0, 0.005)$.
Assume that the true parameter values are $\bar{\beta}_1 = 0$ and $\bar{\beta}_2 = 1$, i.e., the first feature is irrelevant and the second feature is relevant.
Note that the variance is supposed to be known for simplicity.

\subsubsection{Comparison of BM, Lasso, and ARD}

In our setting, we generated 500 datasets, each containing $2 \times 20$ samples of $y$.
Note that, in BM (Algorithm \ref{alg:FAB}), we adopted zero tolerance for model pruning: we pruned the $k$-th feature only when $\hat{\pi}_k$ was smaller than the machine epsilon.
In Lasso, we determined $\alpha$ by 2-fold cross validation.

\begin{figure}[tb]
  \begin{center}
  \includegraphics[width=0.5\textwidth]{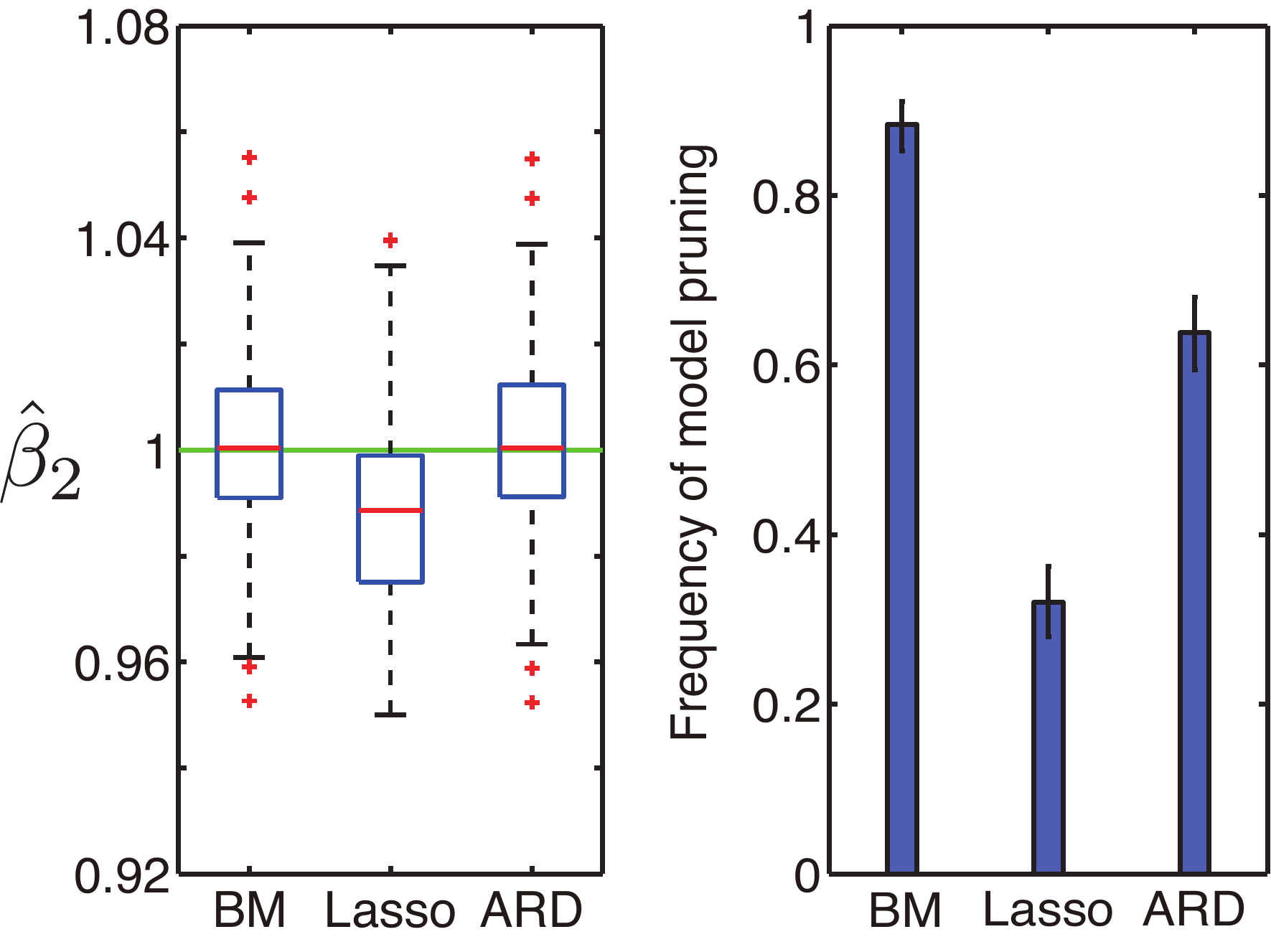}
  \caption{
	Estimation results on synthetic data from Eq.~(\ref{eq:toy_model}).
	(Left) Box plots of estimated values of $\beta_2$. The green line indicates the true value $\bar{\beta}_2 = 1$.
	(Right) Frequency of pruning of the irrelevant feature. The error bars represent the 95\% confidence interval from fitting of the binomial distribution.
  }
  \label{fig:comparison}
  \end{center}
\end{figure}

The estimation results are summarized in Figure~\ref{fig:comparison};
 the left panel shows $\hat{\beta}_2$ when the irrelevant feature was pruned and the right panel shows the frequency of pruning of the irrelevant feature 
 in the 500 trials.  Note that the relevant feature was not pruned in any of the methods.
We can easily see that BM achieved the highest sparsity without shrinkage of $\hat{\beta}_2$.
On the other hand, ARD displayed no visible shrinkage as in BM; however, its sparsity was lower than that of BM.
Lasso displayed shrinkage bias and the lowest sparsity.

\subsubsection{Learning Trajectory of FAB-EM and FAB-EG Algorithms}

Using the same simple example, we show how the parameters are updated by the FAB-EM and FAB-EG algorithms.
For comparison, we also performed the FAB-EG algorithm without reparametrization.
We fixed the learning coefficient as $\eta_t = 2\times10^{-6}$ for FAB-EG and $= 2\times10^{-4}$ for FAB-EG without reparametrization.

\begin{figure}[tb]
  \begin{center}
  \includegraphics[width=0.7\textwidth]{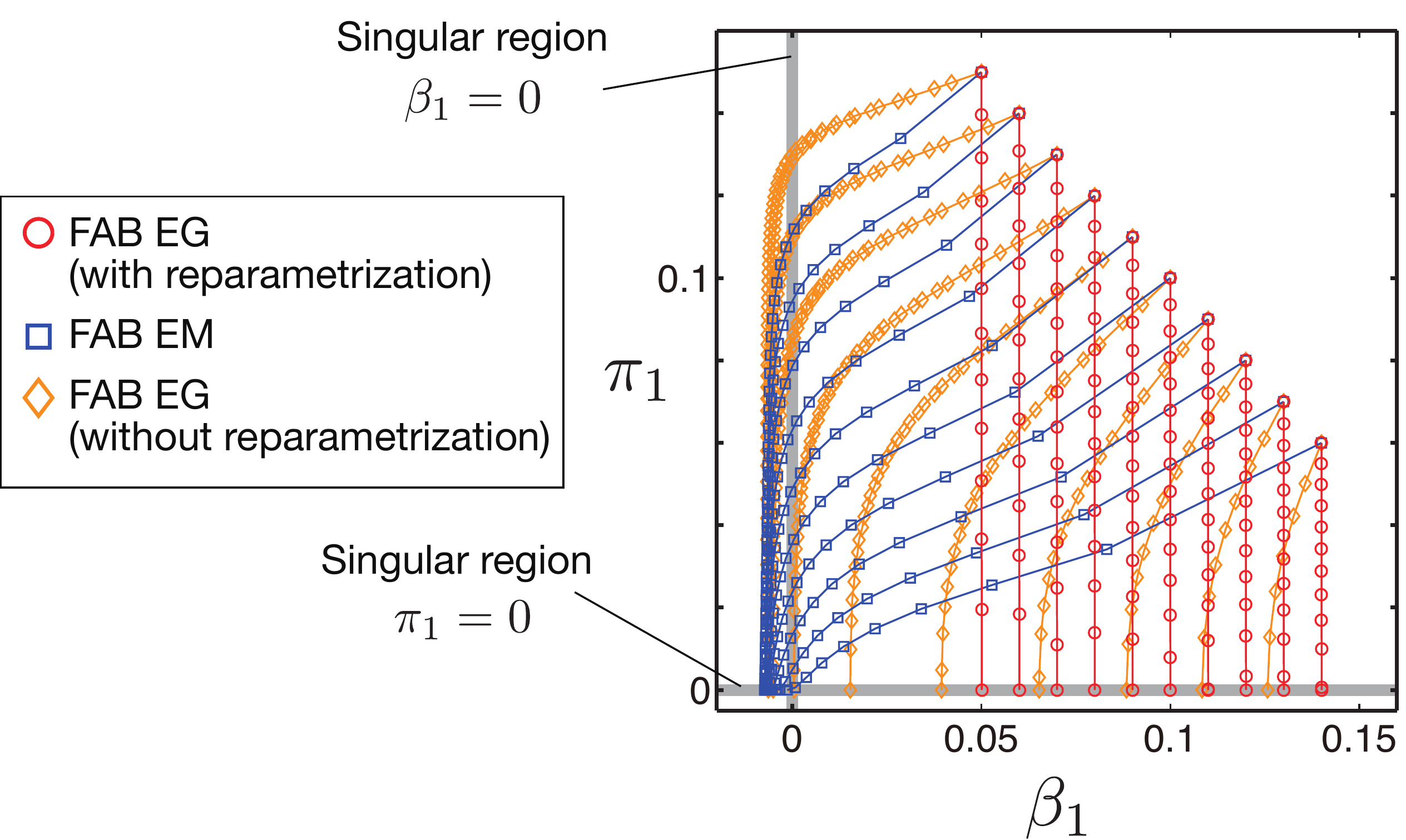}
  \caption{
	Typical learning trajectories on $\beta_1$-$\pi_1$ plane from 10 different initial points
	by FAB-EM steps and FAB-EG steps with/without reparametrization.}
    \label{fig:trajectory}
  \end{center}
\end{figure}

Figure~\ref{fig:trajectory} shows typical learning trajectories of $\beta_1$ and $\pi_1$ with 100 samples.
We considered the 10 initial points of $\beta_1$ and $\pi_1$ located diagonally in the upper right.
The initial values of $\beta_2$ and $\pi_2$ are set to the true values.
In FAB-EM, the learning trajectories were trapped around $\beta_1 = 0$.
In FAB-EG without reparametrization, the trapping was mitigated but still occurred, especially when the initial values of $\beta_1$ were small.
Intuitively speaking, this is because the gradient for $\pi_1$ is shallow with small $\beta_1$.
Thus, the learning trajectory approached to smaller $\beta_1$ since the feature was irrelevant, and then, the learning of $\pi_1$ became slower.
In contrast, the learning trajectories of FAB-EG with the reparametrization approached to $\pi_1 = 0$ with fewer iterations regardless of the initial points,
which means that the irrelevant feature was pruned quickly.
This result empirically demonstrates that using gradient ascent alone improves the convergence only slightly, but combining it with the reparametrization
accelerates the convergence sharply.

\subsection{Experiment with Larger Number of Features}\label{sec:larger_examples}

Next, we explain the results with larger examples ($10 \leq K \leq 100$).
We generated $\bm{\beta}$ and $X$ from the uniform distribution in $[0, 1]$, and half of the elements in $\bm{\beta}$ were set as zero.
$\bm{y}$ was generated by Eq.~\eqref{eq:problem} with $\lambda^{-1} = 0.2$. $N$ was set as $20K$.
We controlled $\{\eta_t\}$ as described in Appendix~\ref{sec:setting_learning_coeff}.

\subsubsection{Performance Validation of Hybrid FAB-EM-EG algorithm}

\begin{figure}[tb]
  \begin{center}
  \includegraphics[width=0.5\textwidth]{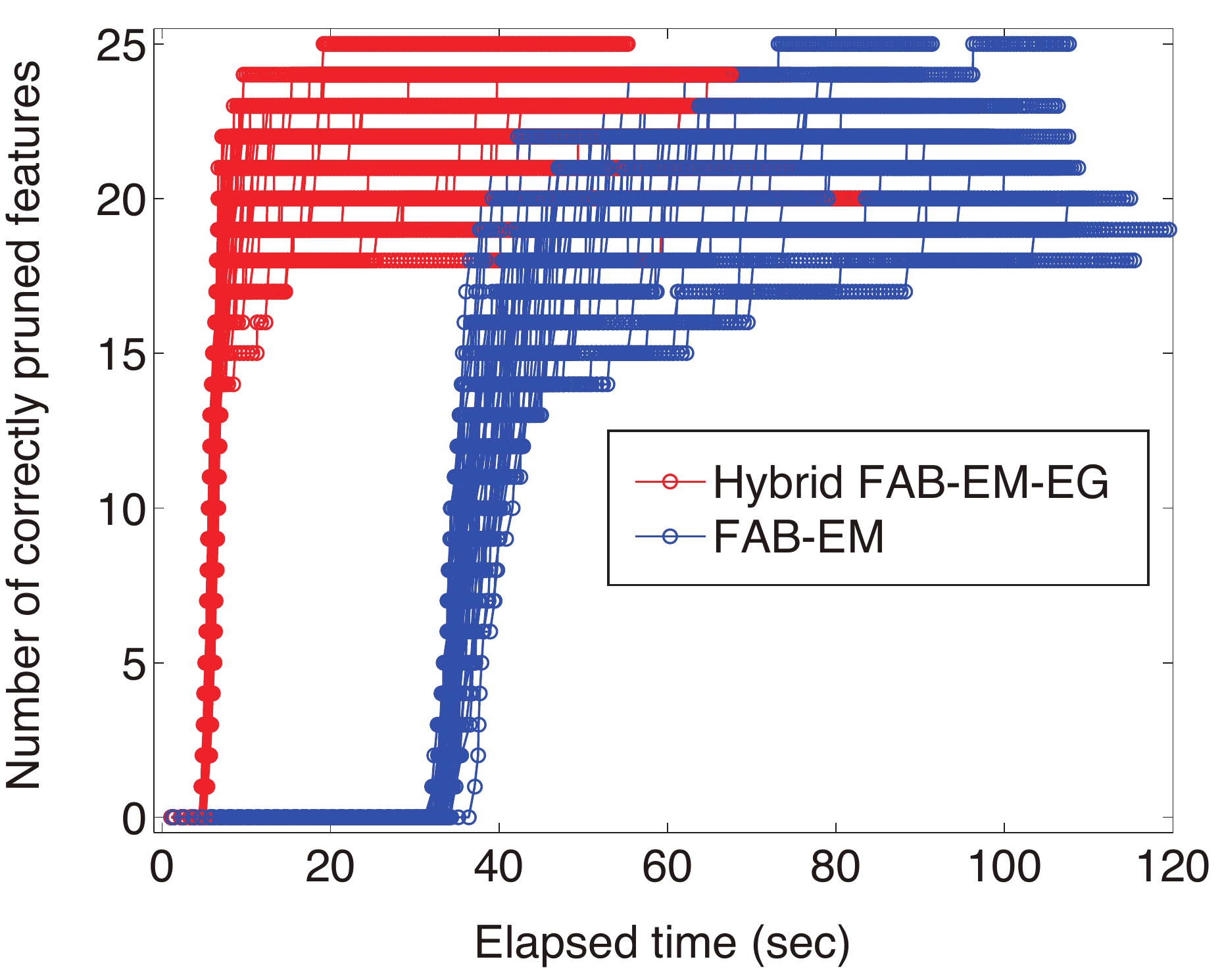}
  \caption{Performance validation of the hybrid FAB-EM-EG algorithm.
  	For the hybrid FAB-EM-EG and the FAB-EM algorithms, the number of correctly pruned features is plotted against the elapsed time.
  	Different lines of the same color represent different datasets.}
    \label{fig:elapsed_time}
  \end{center}
\end{figure}

With $K = 50$, we demonstrate that the hybrid FAB-EM-EG algorithm converges faster than the FAB-EM algorithm.
Toward this end, we counted the number of correctly pruned features and plotted it against the elapsed time for the algorithms.
We set $T$ in Algorithm~\ref{alg:FAB2} to $200$ iterations. Figure~\ref{fig:elapsed_time} shows the number of correctly pruned features against the elapsed time.
We can clearly see the faster convergence of the hybrid FAB-EM-EG algorithm.
Note that the faster convergence is not attributable to over-pruning
because the number of wrongly pruned features at termination were $2.0 \pm 1.3$ (hybrid FAB-EM-EG) and $1.8 \pm 1.3$ (FAB-EM-EG), which were nearly equal.

\subsubsection{Precision and Recall}

\begin{figure}[htp]
  \begin{center}
  \includegraphics[width=0.9\textwidth]{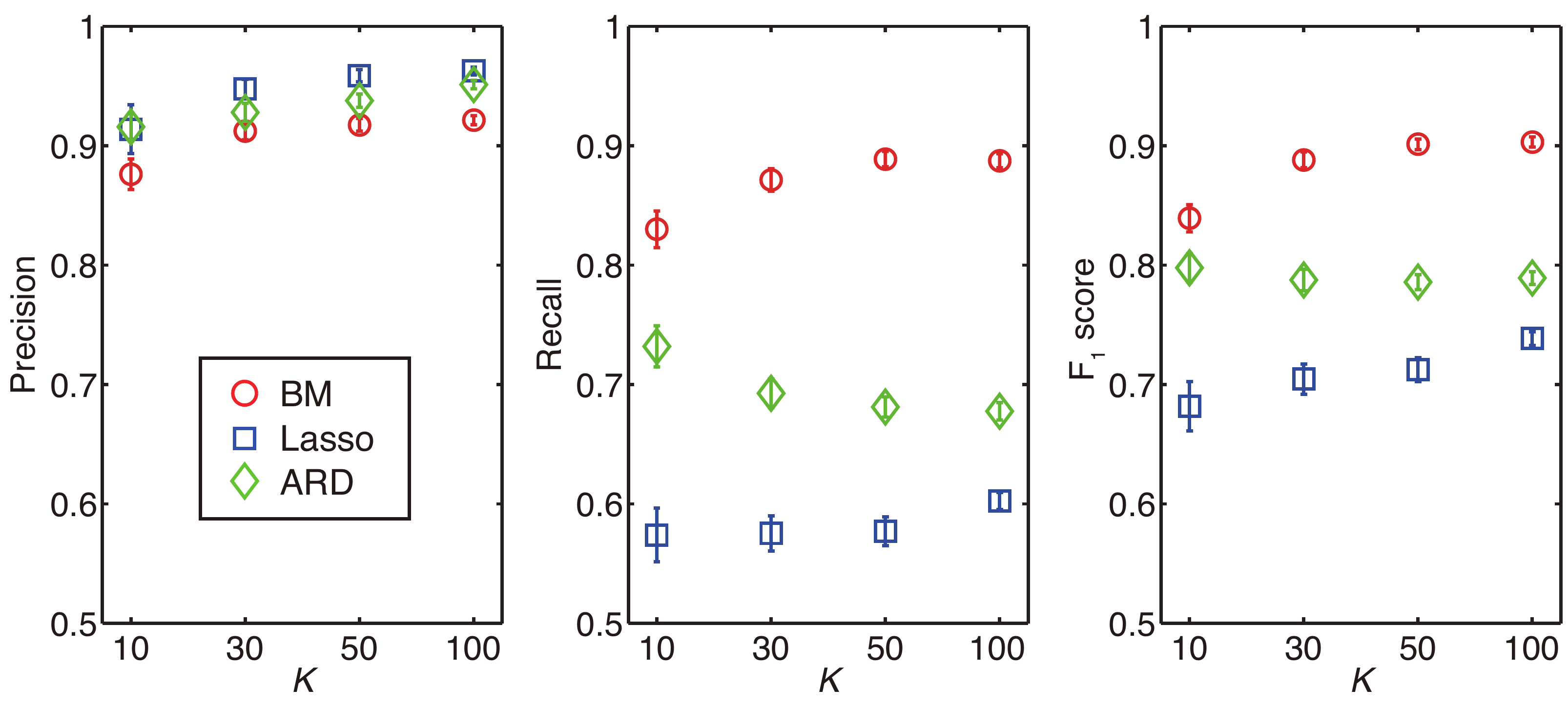}
  \caption{
	Performance measures are plotted with standard errors: (Left) Precision, (Middle) Recall, and (Right) $F_1$ score.}
    \label{fig:comparison2}
  \end{center}
\end{figure}

For $K>2$, we used two performance measures, Recall and Precision, defined as Precision $= m_3/m_2$ and Recall $= m_3/m_1$,
 where $m_1$ and $m_2$ are the numbers of true and estimated irrelevant features, respectively, and $m_3$ is the number of correctly pruned features.
We examined $K = 10$, $30$, $50$, and $100$, and for each $K$, we generated 100 datasets.
Figure~\ref{fig:comparison2} summarizes the estimation results for BM (Algorithm~\ref{alg:FAB2}), Lasso, and ARD.
We set the algorithm switching point $T$ as $500$ iterations. In Lasso, $\alpha$ was determined by 10-fold cross validation.
As shown in the left and middle panels, although BM displayed slightly lower Precision than the others, it achieved the highest Recall.
We also computed the $F_1$ score, the harmonic mean of Precision and Recall, which can be interpreted as a metric for evaluating the performance in terms of the sparsity-shrinkage trade-off.
As shown in the right panel, BM attained the highest $F_1$ score for all $K$ values in this range.
Thus, we concluded that BM achieved the best performance for the larger values of $K$.

%%% Local Variables:
%%% mode: latex
%%% TeX-master: "nips2015.tex"
%%% End:

\section{Conclusion}

In this paper, we proposed a new sparse estimation method, BM, whose key feature is that it does not impose direct regularization on the weights.
Our strategy was to introduce binary latent variables that randomly mask features and to perform a variational Bayesian inference based on FIC.
In addition, we introduced gradient ascent and the reparametrization to accelerate the convergence.
Our analysis of the estimators of BM, Lasso, and ARD highlighted how their sparsity mechanisms are different from one another.
Finally, experimental comparisons of the three methods demonstrated that BM achieves the best performance in terms of the sparsity-shrinkage trade-off.

Note that augmenting a statistical model by random masking variables itself is not a new idea.
For example, \cite{Maaten13} used random masking to generate virtual training data.
However, our approach is distinguished from those studies by its purpose. Namely, we aim to identify whether the features are relevant or not,
rather than improving prediction performance. In the augmented model, the FAB algorithm penalizes the masking rates, i.e., existence probability of the features,
unlike the sparse regularization techniques where the weight values of the features are penalized. Applying the BM to real-world tasks where model identification is crucial,
e.g., causal network inference, is a promising future work.

%%% Local Variables:
%%% mode: latex
%%% TeX-master: "nips2015.tex"
%%% End:

\acks{Y.K. was supported by the Platform Project for Supporting in Drug Discovery and Life Science Research
(Platform for Dynamic Approaches to Living Systems) from Japan Agency for Medical Research and development (AMED).
S.M. and K.H. were supported by JSPS KAKENHI Grant Number 15K15210 and 15K16055, respectively.
}

\bibliography{main}

\appendix
\section{The One-Dimensional ARD Estimator}\label{sec:ard-estimator-when}

According to \cite{wipf08}, the negative marginal log-likelihood when $K=1$ is
given by
\begin{align}
  &\log|\lambda^{-1} I + \gamma\bm{x}\bm{x}^\T|+\bm{y}^\T(\lambda^{-1} I + \gamma\bm{x}\bm{x}^\T)^{-1}\bm{y}
\\
=&\log(\lambda^{-1} + \gamma\bm{x}^\T\bm{x}) + \lambda\bm{y}^\T\bm{y} 
- \bm{y}^\T\frac{\lambda^2\gamma\bm{x}\bm{x}^\T}{1 + \lambda\gamma\bm{x}^\T\bm{x}}\bm{y}
\\
=&\log(\lambda^{-1} + \gamma\bm{x}^\T\bm{x}) + \lambda\bm{y}^\T\bm{y} 
- \frac{\lambda\gamma(\bm{x}^\T\bm{y})^2}{\lambda^{-1} + \gamma\bm{x}^\T\bm{x}}.
\label{eq:mllikelihood}
\end{align}
In the second line, we use the matrix determinant
lemma~\citep[Eq.~(24)]{petersen12} for the first term and the variant
of the Sherman-Morrison relation~\citep[Eq.~(160)]{petersen12} for the second
term.
The derivative is
\begin{align}
  \frac{\partial~\text{Eq.~\eqref{eq:mllikelihood}}}{\partial \gamma}
  &=
  \frac{\bm{x}^\T\bm{x}}{\lambda^{-1} + \gamma\bm{x}^\T\bm{x}}
  - \frac{\lambda(\bm{x}^\T\bm{y})^2}{\lambda^{-1} + \gamma\bm{x}^\T\bm{x}}
  + \frac{\lambda\gamma(\bm{x}^\T\bm{y})^2\bm{x}^\T\bm{x}}{(\lambda^{-1} + \gamma\bm{x}^\T\bm{x})^2}
\\
  &=
  \frac{%
\bm{x}^\T\bm{x}(\lambda^{-1} + \gamma\bm{x}^\T\bm{x}) 
- \lambda(\bm{x}^\T\bm{y})^2(\lambda^{-1} + \gamma\bm{x}^\T\bm{x})
+ \lambda\gamma(\bm{x}^\T\bm{y})^2\bm{x}^\T\bm{x}
}{(\lambda^{-1} + \gamma\bm{x}^\T\bm{x})^2}
\\
  &=
  \frac{%
\lambda^{-1}\bm{x}^\T\bm{x} + \gamma(\bm{x}^\T\bm{x})^2
- (\bm{x}^\T\bm{y})^2
}{(\lambda^{-1} + \gamma\bm{x}^\T\bm{x})^2}.
\end{align}
The stationary point is then given as
\begin{align}\label{eq:gamma-opt}
  \hat\gamma &= \max(0,\frac{(\bm{x}^\T\bm{y})^2 - \lambda^{-1}\bm{x}^\T\bm{x}}{(\bm{x}^\T\bm{x})^2})
\\
&= \max(0, \hat{\beta}_{\rm{LS}}^2 - (\lambda\bm{x}^\T\bm{x})^{-1}).
\end{align}
Note that we use the max operator since $\gamma$ is the variance and
it must be non-negative.
By substituting the result of $\hat\gamma>0$ into Eq.~\eqref{eq:ard}, we
obtain
\begin{align}  
  \hat{\beta}_{\rm{ARD}} 
  &= \hat\gamma\bm{x}^\T(\lambda^{-1} I + \hat\gamma\bm{x}\bm{x}^\T)^{-1}\bm{y}
\\
&= \lambda\hat\gamma\bm{x}^\T\bm{y} 
- \frac{\lambda\hat\gamma^2\bm{x}^\T\bm{x}\bm{x}^\T\bm{y}}{\lambda^{-1} + \hat\gamma\bm{x}^\T\bm{x}}
\\
&= \lambda\hat{\beta}_{\rm{LS}}^2 \bm{x}^\T\bm{y} - \hat{\beta}_{\rm{LS}}
- \frac{%
  \lambda\hat{\beta}_{\rm{LS}}^4\bm{x}^\T\bm{x}\bm{x}^\T\bm{y}
  -2 \hat{\beta}_{\rm{LS}}^2\bm{x}^\T\bm{y}
  + \frac{\lambda\bm{x}^\T\bm{y}}{\lambda^2\bm{x}^\T\bm{x}}
}{\lambda^{-1} + \hat{\beta}_{\rm{LS}}^2\bm{x}^\T\bm{x} - \lambda^{-1}}
\\
&= \lambda\hat{\beta}_{\rm{LS}}^2 \bm{x}^\T\bm{y} - \hat{\beta}_{\rm{LS}}
- \frac{%
  \lambda\hat{\beta}_{\rm{LS}}^4\bm{x}^\T\bm{x}\bm{x}^\T\bm{y}
  -2 \hat{\beta}_{\rm{LS}}^2\bm{x}^\T\bm{y}
  + \lambda^{-1}\hat{\beta}_{\rm{LS}}
}{\hat{\beta}_{\rm{LS}}^2\bm{x}^\T\bm{x}}
\\
&= - \hat{\beta}_{\rm{LS}}
+2 \hat{\beta}_{\rm{LS}}
- \frac{1}{\lambda\hat{\beta}_{\rm{LS}}\bm{x}^\T\bm{x}}
\\
&= \hat{\beta}_{\rm{LS}}- \frac{1}{\lambda\bm{x}^\T\bm{y}}.
\end{align}
Recall that $\hat{\beta}_{\rm{ARD}}=0$ when $\hat\gamma\leq 0$. Since
$\hat{\beta}_{\rm{LS}}^2$ and $(\lambda\bm{x}^\T\bm{x})^{-1}$ are both
non-negative, the condition $\hat\gamma\leq 0$ is written as
$\hat{\beta}_{\rm{LS}}^2 \leq (\lambda\bm{x}^\T\bm{x})^{-1}$, or
equivalently, $|\hat{\beta}_{\rm{LS}}| \leq
|\lambda\bm{x}^\T\bm{y}|^{-1}$.
Substituting this
condition into the above equation yields Eq.~\eqref{eq:ard-1d}.
%

%%% Local Variables:
%%% mode: latex
%%% TeX-master: "acml15_ard_debug.tex"
%%% End:

\section{Derivation of The Lower Bound of FIC}\label{sec:derivation_lowerbound}

\cite{hayashi15} obtained a general representation of the lower bound of FIC, and in this case, we have
\begin{eqnarray}
	\mbox{FIC} \geq \kitai_q [\log p(\bm{y},Z | X,\bm{\beta}, \lambda, \bm{\pi})] - \frac{1}{2} \kitai_q [\log | F_{\bm{\beta}} |] - \frac{K+1}{2} \log N + H(q),
\end{eqnarray}
where $q = q(Z)$ and $F_{\bm{\beta}}$ denotes the Hessian matrix of the negative log-likelihood w.r.t. $\bm{\beta}$.
Note that the priors for $\bm{\beta}$ and $\lambda$ do not appear in the lower bound, since the priors do not depend on $N$, as assumed in Section~\ref{sec:bm_model}.
In order to derive the FAB algorithm, we compute the lower bound of the second term on the right-hand side.

The Hessian matrix $F_{\bm{\beta}}$ is represented as
\begin{eqnarray}
	F_{\bm{\beta}} &=& - \nabla_{\bm{\beta}} \nabla_{\bm{\beta}} \log p(\bm{y},Z | X,\bm{\beta}, \lambda, \bm{\pi})\\
	&=& (X \circ Z)^\top (X \circ Z).
\end{eqnarray}
Then, using Hadamard's inequality for a positive-semidefinite matrix yields
\begin{eqnarray}
	-\log |F_{\bm{\beta}}| &\geq& -\sum_k \log \sum_n x_{nk}^2 z_{nk}\\
	&\geq& - \sum_k \log (\sum_n z_{nk}) (\sum_n x_{nk}^2)\\
	 &=& - \sum_k \log \sum_n z_{nk} + \mbox{const}.
\end{eqnarray}
As stated in \cite{fujimaki12a}, since $- \log \sum_n z_{nk}$ is a concave function, its linear approximation at $N\pi_k > 0$ yields the lower bound:
\begin{eqnarray}
	-\kitai_q [\log \sum_n z_{nk}] \geq - \left( \log N\pi_k + \frac{\sum_n \kitai_q [z_{nk}] /N - \pi_k}{\pi_k}\right).
\end{eqnarray}
Thus, we obtain Eq.~(\ref{eq:FIClowerbound}) in the main text.

%%% Local Variables:
%%% mode: latex
%%% TeX-master: "acml15_ard_debug.tex"
%%% End:

\section{Control of Learning Coefficient}\label{sec:setting_learning_coeff}

We explain how to set the learning coefficients $\{\eta_t\}$ in Section~\ref{sec:larger_examples}. We set $\eta_t$ to a constant value $\eta$; however, when the maximum of the update of $\{\pi_k\}$ is greater than $0.05$, $\eta_t$ is modified such that the maximum is $0.05$.
We chose the constant $\eta$ as $2 \times 10^{-2} / N$, where $N$ is the number of samples.

%%% Local Variables:
%%% mode: latex
%%% TeX-master: "acml15_ard_debug.tex"
%%% End:

\end{document}